\pdfoutput=1
\documentclass[11pt]{article}

\usepackage[preprint]{acl}

\usepackage{times}
\usepackage{latexsym}

\usepackage[T1]{fontenc}
\usepackage[utf8]{inputenc}

\usepackage{microtype}

\usepackage{inconsolata}

\usepackage{graphicx}

\usepackage{microtype}
\usepackage{graphicx}
\usepackage{subfigure}
\usepackage{amsmath}
\usepackage{amsfonts}
\usepackage{multirow}
\usepackage{color}
\usepackage{amssymb}
\usepackage{booktabs}
\usepackage{arydshln}
\usepackage{algorithm}
\usepackage{algorithmic}
\usepackage{pifont}
\title{ListConRanker: A Contrastive Text Reranker with Listwise Encoding}

\author{
 \textbf{Junlong Liu\textsuperscript{1,2}\footnotemark[1]},
 \textbf{Yue Ma\textsuperscript{2}},
 \textbf{Ruihui Zhao\textsuperscript{2}},
 \textbf{Junhao Zheng\textsuperscript{1}},
 \textbf{Qianli Ma\textsuperscript{1}\footnotemark[2]},
 \textbf{Yangyang Kang\textsuperscript{2,3}\footnotemark[2]}
\\
\\
 \textsuperscript{1}School of Computer Science and Engineering, South China University of Technology \\
 \textsuperscript{2}ByteDance China,
 \textsuperscript{3}Zhejiang University
\\
    \texttt{junlongliucs@foxmail.com}, \texttt{junhaozheng47@outlook.com}, \texttt{qianlima@scut.edu.cn} \\
    \texttt{\{mayue.ian, zhaoruihui, yangyangkang\}@bytedance.com}
}

\begin{document}
\maketitle

\renewcommand{\thefootnote}{\fnsymbol{footnote}}
\footnotetext[1]{This work was conducted when Junlong Liu was interning at ByteDance.}
\footnotetext[2]{Corresponding authors}

\renewcommand{\thefootnote}{\arabic{footnote}}

\begin{abstract}
Reranker models aim to re-rank the passages based on the semantics similarity between the given query and passages, which have recently received more attention due to the wide application of the Retrieval-Augmented Generation. Most previous methods apply pointwise encoding, meaning that it can only encode the context of the query for each passage input into the model. However, for the reranker model, given a query, the comparison results between passages are even more important, which is called listwise encoding. Besides, previous models are trained using the cross-entropy loss function, which leads to issues of unsmooth gradient changes during training and low training efficiency. To address these issues, we propose a novel \textbf{List}wise-encoded \textbf{Con}trastive text re\textbf{Ranker} (\textbf{ListConRanker}). It can help the passage to be compared with other passages during the encoding process, and enhance the contrastive information between positive examples and between positive and negative examples. At the same time, we use the circle loss to train the model to increase the flexibility of gradients and solve the problem of training efficiency. Experimental results show that ListConRanker achieves state-of-the-art performance on the reranking benchmark of Chinese Massive Text Embedding Benchmark, including the cMedQA1.0, cMedQA2.0, MMarcoReranking, and T2Reranking datasets.\footnote{Our codes and weights are publicly available at \url{https://huggingface.co/ByteDance/ListConRanker}}
\end{abstract}

\section{Introduction}
Given a query and a few passages that are semantically related or partially related to the query, the goal of the ranking task is to sort these passages based on their degree of semantic similarity. The reranker model is one of the most important parts in the Information Retrieval (IR) systems \citep{GUO2020102067, 10.1145/3397271.3401104, 10.1007/978-3-030-45442-5_37}. Recently, due to the trend of Retrieval-Augmented Generation (RAG) and limited context length of Large Language Models (LLMs), reranking models have been widely studied and rapidly developed \citep{10.5555/3495724.3496517, zhao2024retrieval, glass-etal-2022-re2g}, which can filter out noisy passages retrieved by embedding models.

\begin{figure}[t!]
    \centering
    \includegraphics[width=\linewidth]{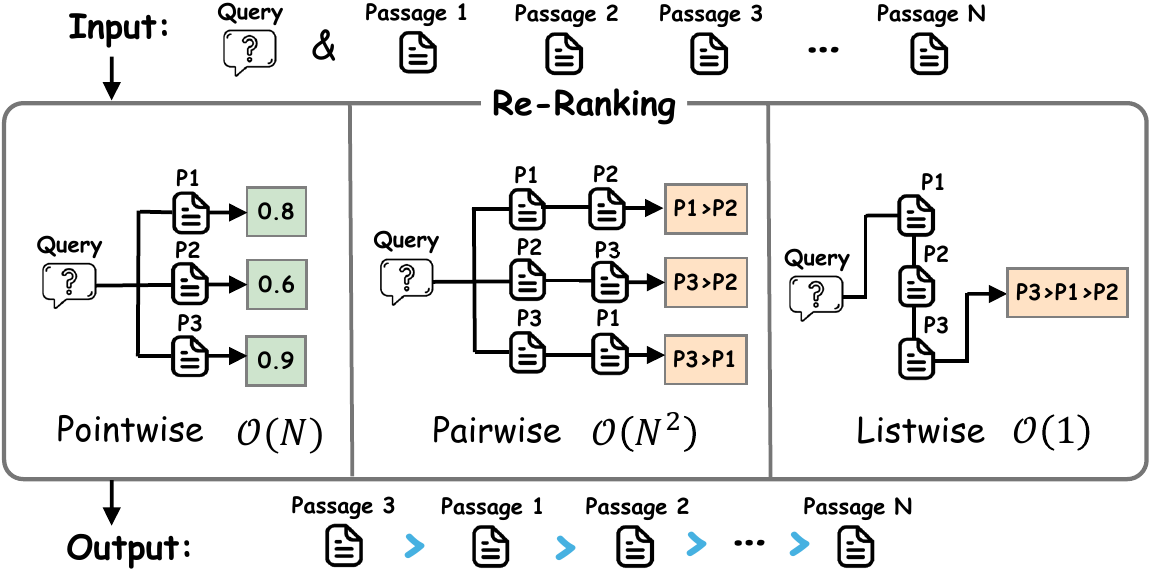}
    \caption{
    Pointwise rerankers receive the query and a passage as input and output the similarity score of them. Pairwise rerankers receive the query and two passages as input and output which passage is more similar to the query. Listwise rerankers receive the query and all passages as input and directly output the final ranking results.
    }
    \label{fig:example}
\end{figure}

As shown in Figure~\ref{fig:example}, there are three main types of reranking models, which are pointwise \citep{nogueira-etal-2020-document, liang2023holistic, sachan-etal-2022-improving}, pairwise \citep{qin-etal-2024-large}, and listwise \citep{sun-etal-2023-chatgpt, ma2023zero}. The main difference between them is that a pointwise reranker accepts a query and one passage as input at a time, while a pairwise reranker takes two passages, and a listwise reranker takes multiple passages at a time.

Most previous rerankers based on BERT-like Cross-Encoder are pointwise \citep{lu2022ernie}, because of the limited context length and the need to align with the two-sentence input format used during pre-training \citep{devlin-etal-2019-bert, zhuang-etal-2021-robustly}. However, pointwise rerankers can only obtain the context of the query and a passage, when models encode the feature of passage. The most important aspects are the relative ordering of passages and the comparison of similarities between passages. This results in the suboptimal performance of pointwise rerankers. Pairwise and listwise encoding reranker are two solutions to this problem. Due to the high computational complexity 
and unstable comparison results (e.g., passage 1 > 2 and passage 2 > 3, but passage 3 > 1) of pairwise methods, their practical application value is low. 
% Previously, only a few studies have explored this approach. 
In contrast, listwise methods represent a more realistic and valuable direction for exploration.

Recently, LLMs have demonstrated excellent performance on natural language tasks. Some work has attempted to use LLMs as listwise rerankers \citep{pradeep2023rankvicuna, liu2024leveraging}. They allow the LLM to directly output the predefined sequence numbers of passages as the final ranking result. However, it will still exceed the context length limit when there are hundreds of passages, leading to the failure of ranking \citep{sun-etal-2023-chatgpt}. Besides, LLMs can not output exact similarity scores between the query and passages. Moreover, different input orders may also lead to instability in the results.

Based on the problems of the pointwise reranker and the listwise reranker based on LLMs, we explore the listwise reranker based on BERT-like embedding models, which is an area that has not been fully explored before. We propose a \textbf{List}wise-encoded \textbf{Con}trastive text re\textbf{Ranker} (\textbf{ListConRanker}). It first inputs query and passages into the embedding model. After getting the original features of the query and each passage, we combine these features into an input sequence, which is then fed into the proposed ListTransformer. The ListTransformer can facilitate global contrastive information learning between passage features, including the clustering of similar passages, the clustering between dissimilar passages, and the distinction between similar and dissimilar passages. Besides, we propose ListAttention to help ListTransformer maintain the features of the query while learning global comparative information.

At the same time, most of the previous pointwise rerankers apply cross-entropy loss function during training. However, for the ranking task, there is usually more than one similar (positive) passage. If the cross-entropy loss function is used, for multiple passages of the same query, models can only sample one positive passage per training step. This leads to low training efficiency for positive samples. Besides, it hinders the learning of contrastive information between positive samples.

To this end, we propose to apply Circle Loss \citep{9156774}. It can improve the data efficiency by taking multiple positive and negative passages as input at the same time.
Besides, its self-adaptive weights can smooth out gradient changes during the training process and accelerate training. Specifically, it can reduce the gradient when close to the optimal space while increasing the gradient when far from it. This helps the model find the optimal space more quickly.

The main contributions of our work can be summarized as follows:
\begin{itemize}
    \item We propose a novel BERT-based reranker model with listwise encoding named ListConRanker, which has not been fully explored before. It contains ListAttention and ListTransformer, which can utilize the global contrastive information to learn representations.
    \item We propose to use the Circle Loss as the loss function. Compared with cross-entropy loss and ranking loss, it can solve the problems of low data efficiency and unsmooth gradient change. 
    % Besides, the training objective of Circle Loss is consistent with the task objective.
    \item Experimental results on the reranking benchmark of Chinese Massive Text Embedding Benchmark (C-MTEB) demonstrate that ListConRanker is state-of-the-art. The ablation study shows the effectiveness of ListTransformer and Circle Loss.
\end{itemize}

\section{Related Work}

There are three main encoding types of rerankers: pointwise encoding, pairwise encoding, and listwise encoding. As shown in Figure~\ref{fig:example}, the main differences between them are the number of passages input at a time and the output format. Due to the high computational complexity of pairwise rerankers, which is $O(N^2)$ for ranking $N$ passages, there has been less related work in the past. In the following sections, we will mainly introduce these three rerankers.

\subsection{Pointwise Rerankers}

Pointwise rerankers receive the query and a passage as input, and output the similarity between them. \citet{10.1145/3331184.3331303} and \citet{10.1007/978-3-030-72240-1_26} were the first to propose using the BERT model in ranking tasks and referred to this type of model as the cross-encoder architecture. This allows for interaction between the query and the passage. However, they used the binary cross-entropy loss function during training, which prevented interaction between query-passage pairs. Based on this, \citet{nogueira2019passage} proposed using contrastive loss from a new perspective. Recently, due to the huge success of generative language models, \citet{nogueira-etal-2020-document} proposed to input the query and passage into the model directly and then let the model generate a "true" or "false" token to indicate whether they are similar. Furthermore, \citet{sachan-etal-2022-improving} proposed to use LLMs to compute the probability of the input query conditioned on the passage as the similarity between them.

However, these pointwise rerankers can only learn the context of one passage, which results in suboptimal performance. Listwise rerankers can provide global contrastive information, making it superior in architecture.

\subsection{Pairwise and Listwise Rerankers}
Pairwise Rerankers take a query and two passages as input and output which of the two passages is more similar to the query. \citet{qin-etal-2024-large} proposed using an LLM as a judge. After obtaining results for all passage pairs, a traditional ranking algorithm is used to produce the final passage ranking.

Listwise rerankers receive the query and all corresponding passages as input. It outputs the ranking of passages directly. Most previous methods are based on LLMs. For example, some methods first proposed to let LLMs generate the reordered list \citep{ma2023zero, pradeep2023rankvicuna, zhang2023rank}. To address the issue of exceeding the context length of LLMs in some extreme cases, \citet{sun-etal-2023-chatgpt} proposed a listwise method based on sliding windows. Furthermore, \citet{liu2024leveraging} encoded the passage into an embedding before inputting it into the LLM. This helps to address the lack of global contrastive information in the methods based on sliding windows. 

Due to the limited context length of LLMs, the rerankers based on LLMs can not provide complete global contrastive information, which is the most important advantage of listwise rerankers.

\begin{figure*}[t!]
    \centering
    \includegraphics[width=\linewidth]{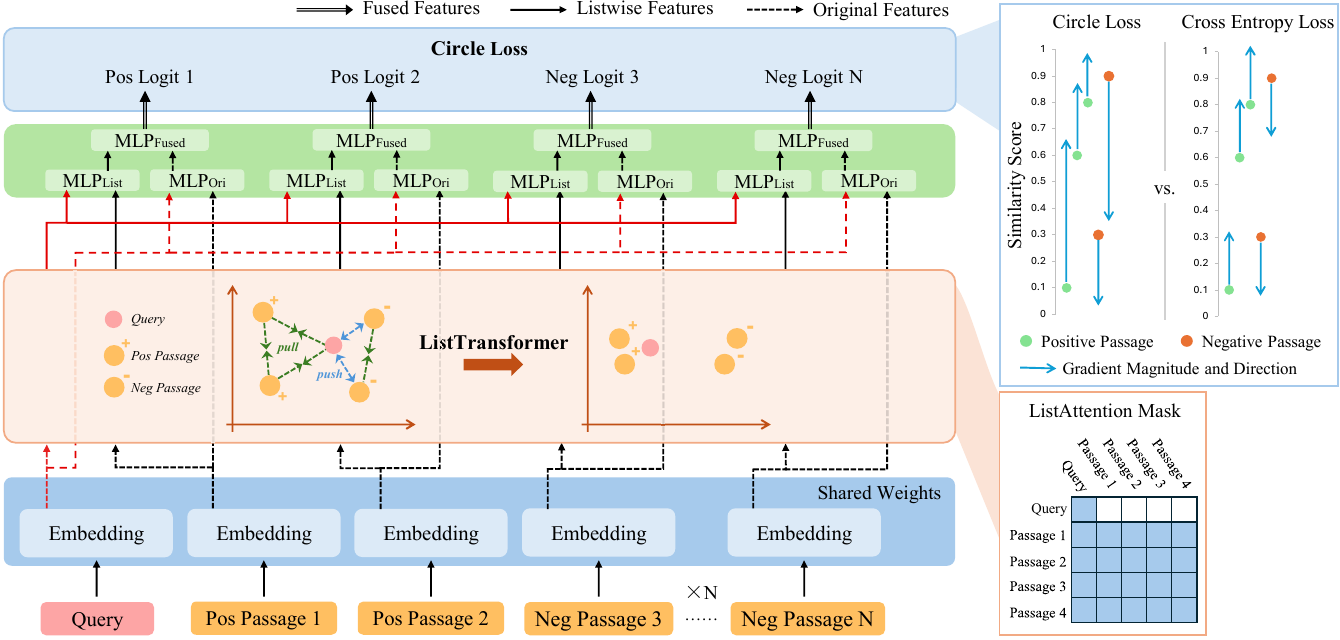}
    \caption{
    The overall architecture of the ListConRanker.
    }
    \label{fig:method}
\end{figure*}

\section{Method}
In this section, we mainly describe ListConRanker, which applies listwise encoding to learn global contrastive information. It uses the Circle Loss \citep{9156774} to improve the data efficiency and smooth out gradient changes. The structure of ListConRanker is shown in Figure~\ref{fig:method}.

\subsection{Original Features}
Given a query $q$ and $N$ passages $P=\{p_1, p_2, ... ,p_N\}$, we first input them to an embedding model to obtain their features $H=\{h_q, h_1, h_2, ..., h_N\}$.

\begin{gather}
    h_q = \text{Embedding}(q) \\
    h_i = \text{Embedding}(p_i)
\end{gather}
where $\text{Embedding}$ is a BERT-based embedding model, $h_q$ is the feature of query, and $h_i$ is the feature of passage $i$.

We get the original features by using an embedding to encode the context of the query and passage. However, these features do not contain any information of other passages, which means these passages are unable to be compared with other passages to optimize the ranking result.

\subsection{Listwise Encoding by ListTransformer and ListAttention}
After obtaining the features of query and passages, we concatenate these features into a sequence and input them into the ListTransformer which is similar to a Transformer Encoder \citep{vaswani2017attention} except for the self-attention module. In addition, we apply learnable embeddings to help ListTransformer distinguish between query feature and passage features.

\begin{gather}
    Z^{(0)} = \{h_q+e_q, h_1+e_p, ..., h_n+e_p\} \\
    Q^{(l)} = Z^{(l - 1)}W_Q^{(l)} \\
    K^{(l)}=Z^{(l - 1)}W_K^{(l)} \\
    V^{(l)}=Z^{(l - 1)}W_V^{(l)} \\
    Z^{(l)} = \text{ListAttention}(Q^{(l)}, K^{(l)}, V^{(l)})
\end{gather}
where $e_q$ and $e_p$ are learnable embeddings, $W_Q^{(l)}$, $W_K^{(l)}$, and $W_V^{(l)}$ are learnable parameters, and $l$ is the $l$-th layer of ListTransformer.

The bidirectional self-attention in ListTransformer will make the query feature obscured by unrelated passages when there are a large number of unrelated passages. When the number of unrelated passages is small, the query can ignore the unrelated passages by attention weights. However, when the number is large, the noisy information dominates self-attention.

To this end, we propose the ListAttention in the ListTransformer. The attention mask details of ListAttention are shown in Figure~\ref{fig:method}. Specifically, we make the query can only get its own attention to preserve its original features. As for passages, we use the bidirectional self-attention. Passages can recognize the query features by query embedding and learn pair features between query and passage. Besides, the attention among passages can facilitate the learning of global contrastive information between passages. For example, it can reduce the distance between positive examples and between negative examples while increasing the distance between positive and negative examples. This global contrastive information reflects the concept of sequencing in ranking tasks.

After obtaining the features with global contrastive information (i.e., listwise features), we combine the listwise features with original features and use MLPs to determine the final similarity between the query and passage.

\begin{gather}
    s^{origin}_i = \text{MLP}_{\text{Ori}}(h_q, h_i)
\end{gather}
\begin{gather}
    s^{list}_i = \text{MLP}_{\text{List}}(z^{(l)}_q, z^{(l)}_i) \\
    s^{final}_i = \sigma (\text{MLP}_{\text{Fused}}(s^{origin}_i, s^{list}_i))
\end{gather}
where $S^{final}=\{s_1^{final}, s_2^{final}, ..., s_n^{final}\}$ is the similarity between the query and passage $p_i$, $\text{MLP}_{\text{Ori}}$, $\text{MLP}_{\text{List}}$, and $\text{MLP}_{\text{Fused}}$ are all MLPs, $\sigma$ is the sigmoid activation function.

Finally, we can sort all passages by their similarity to the query $S^{final}$ in descending order to obtain the final ranking result.

\subsection{Circle Loss}

Most of the previous rerankers apply cross-entropy loss function to train the model. However, cross-entropy loss function has several disadvantages. First, its data efficiency is low. It can only sample one similar (positive) passage from all the passages per training step. This also prevents global contrast information learning between positive examples. Second, it does not have the feature of adaptively adjusting loss weights. Specifically, when the predicted values are close to or far from having the optimal space, the cross-entropy loss function can only implicitly control the size of parameter updates through the magnitude of the gradient. At the same time, the ordering is a relative result in ranking tasks. It only requires that positive samples be ranked higher than negative samples. The cross-entropy loss function requires the predictions for positive cases to be 1 and for negative cases to be 0. This is too strict for ranking tasks.

Therefore, we propose using Circle Loss\citep{9156774} as the loss function for ListConRanker. It can customize the optimal prediction values for positive and negative cases. At the same time, it can adaptively adjust the gradient weights for each sample based on the customized optimal prediction values, which helps the model achieve smoother gradient changes. Ultimately, this makes it easier for the model to find the optimal space. The calculation method for Circle Loss is as follows:

\begin{gather}
    L = \text{log} (1+ r_{neg} r_{pos}) \\
    r_{neg} = \sum^J_{j=1} \text{exp} (\gamma \alpha_{neg}^j(s_{neg}^j - \Delta_{neg})) \\
    r_{pos} = \sum^I_{i=1} \text{exp} (-\gamma \alpha^i_{pos}(s^i_{pos}-\Delta_{pos})) \\
    \Delta_{neg} = m, \, \Delta_{pos} = 1-m \\
    O_{neg} = -m, \, O_{pos} = 1 + m \\
    \alpha_{neg}^j = \text{max}(0, s^j_{neg} - O_{neg}) \\
    \alpha^i_{pos} = \text{max}(0, O_{pos} - s^i_{pos})
\end{gather}
where $m$ and $\gamma$ are hyper-parameters, $I$ is the number of positive samples in all passages, $J$ is the number of negative samples in all passages.

\subsection{Iterative Inference}
Due to the limited memory of GPUs, we input about 20 passages at a time for each query during training. However, during actual use, there may be situations where far more than 20 passages are input at the same time. For example, each query corresponds to approximately 1,000 passages that need to be ranked in the MMarcoReranking \citep{bonifacio2021mmarco} test set. The differences between the training and inference processes can cause self-attention weights to become dispersed in ListTransformer, which prevents the passages from learning effective global contrastive information. This has been fully demonstrated in the context length extension problem of large language models \citep{chen2023extending, alibi}. Finally, it leads to a drop in performance.

However, we find that ListConRanker still maintains a basic ranking capability when a large number of passages are input simultaneously through case studies. Specifically, for some clearly dissimilar passages, ListConRanker can still rank them after partially similar samples (i.e., hard negative samples) or similar samples (positive samples). When a large number of passages are input, ListConRanker shows insufficient discrimination ability between partially similar samples or similar samples. On the contrary, when only dozens of passages are input, ListConRanker can easily distinguish between partially similar passages and similar samples through global contrastive information.

\begin{algorithm}[t]
    \caption{Iterative Inference}
    \label{alg:iterative}
    \renewcommand{\algorithmicrequire}{\textbf{Input:}}
    \renewcommand{\algorithmicensure}{\textbf{Output:}}
    \begin{algorithmic}[1]
        \REQUIRE Query $q$, Passage $P=\{p_1,p_2,...,p_N\}$, Termination condition of the iteration $\alpha$, Passage reduction rate per iteration $\beta$  %%input
        \ENSURE Ranked result $R=\{r_1, r_2, ..., r_N\}$, where $r_i$ is the order of $p_i$   %%output
        \WHILE{$|P|>\alpha$}
            \STATE $nums \gets \text{ceil}(|P| \times \beta)$
            \STATE $scores \gets \text{ListConRanker}(q,P)$
            \STATE $orders \gets \text{argsort}(scroes)$  \text{// Descending}
            \FOR{each $i \in [1,nums]$}
		      \STATE $last\_one \gets orders[-i]$
                \STATE $r_{last\_one} \gets |P|$
                \STATE $P.\text{remove}(p_{last\_one})$
            \ENDFOR
        \ENDWHILE
        
        \STATE $scores \gets \text{ListConRanker}(q,P)$
        \STATE $orders \gets \text{argsort}(scroes)$ \text{// Descending}
        \FOR{each $i \in [1,|P|]$}
		\STATE $last\_one \gets orders[-i]$
            \STATE $r_{last\_one} \gets |P|$
            \STATE $P\text{.remove}(p_{last\_one})$
        \ENDFOR

        \RETURN $R$
    \end{algorithmic}
\end{algorithm}

\begin{figure}
    \centering
    \includegraphics[width=\linewidth]{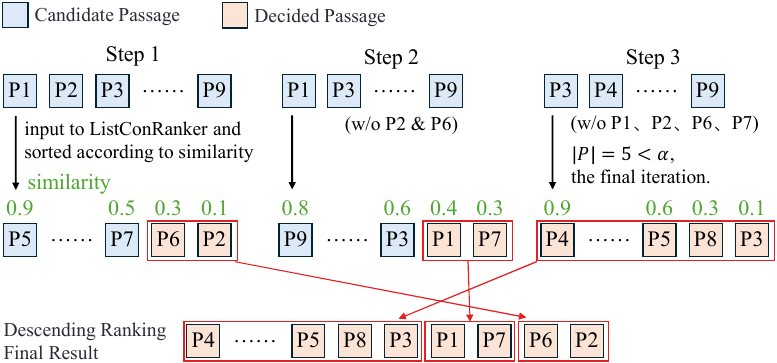}
    \caption{
    The process of iterative inference. For simplicity, we do not show the query in the figure. The query will be input during each step of the inference.
    }
    \label{fig:iterative}
\end{figure}

Based on the above findings, we propose a novel iterative inference method for ListConRanker. The specific inference process is shown in Algorithm~\ref{alg:iterative} and Figure~\ref{fig:iterative}. For each inference step $i$, we input all $N_i$ passages and then determine the ranking positions for the bottom $( \beta \times N_i )$ passages in the final results. The remaining $ N_{i+1}=(1-\beta) \times N_i$ passages are re-input into the ListConRanker. The iteration continues until the number of remaining passages is less than $\alpha$. The final positions for all passages are obtained.

\begin{table*}[t]
\centering
\resizebox{2 \columnwidth}{!}{
\begin{tabular}{lcccccc}
\toprule
Model                   &LLM?  & cMedQA1.0 & cMedQA2.0 & MMarcoReranking & T2Reranking & Avg.  \\ \hline
% gte-Qwen2-7B-instruct     & 88.20     & 88.03     & 31.65           & 67.80       & 68.92 \\
360Zhinao-search    & \ding{56}      & 87.00     & 88.48     & 32.41           & 67.80       & 68.92 \\
% IYun-large-zh           & \ding{56}  & 88.74     & 89.42     & 31.61           & 67.45       & 69.30 \\
Baichuan-text-embedding  & \ding{56} & 88.06     & 88.46     & 34.30           & 67.85       & 69.67 \\
Yinka                   & \ding{56}  & 89.26     & 90.05     & 32.74           & 67.05       & 69.78 \\
piccolo-large-zh-v2     & \ding{56}  & 89.31     & 90.14     & 33.39           & 67.15       & 70.00 \\
360Zhinao-1.8B-Reranking  & \ding{52} & 86.75     & 87.92     & 37.29           & 68.55       & 70.13 \\
ternary-weight-embedding  & \ding{56} & 88.74     & 88.38     & 35.04           & 68.39       & 70.14 \\
LdIR-Qwen2-reranker-1.5B  & \ding{52} & 86.50     & 87.11     & 39.35           & 68.84       & 70.45 \\
zpoint-large-embedding-zh & \ding{56} & 91.11     & 90.07     & 38.87           & \textbf{69.29}       & 72.34 \\
xiaobu-embedding-v2     & \ding{56}  & 90.96     & \textbf{90.41}     & 39.91           & 69.03       & 72.58 \\
Conan-embedding-v1     & \ding{56}   & \textbf{91.39}     & 89.72     & 41.58          & 68.36       & 72.76 \\ \hline
ListConRanker        & \ding{56}    & 90.55     & 89.38     & \textbf{43.88}           & 69.17       & \textbf{73.25} \\
- w/o Iterative Inference & & 90.19     & 89.93     & 37.52           & 69.17       & 71.70 \\
\bottomrule
\end{tabular}
}
\caption{The results comparison with baselines on the reranking benchmark of C-MTEB. We select the top 10 open-source models or models providing API inference as of December 10, 2024, based on the average metrics across cMedQA1.0, cMedQA2.0, MMarcoReranking, and T2Reranking as the baselines. The evaluation metric for all datasets is mAP. The best performance is in \textbf{bold}.
}
\label{tab:overall results}
\end{table*}

\section{Experiments}

\subsection{Dataset and Evaluation Metrics}
We evaluate ListConRanker on the reranker section of Chinese Massive Text Embedding Benchmark (C-MTEB)\footnote{\url{https://huggingface.co/spaces/mteb/leaderboard}} \citep{10.1145/3626772.3657878}, including the test set of cMedQA1.0 \citep{app7080767}, test set of cMedQA2.0 \citep{8548603}, development set of MMarcoReranking, and development set of T2Reranking \citep{10.1145/3539618.3591874}. We report the mean Average Precision (mAP) on each dataset to show the effectiveness of reranker models.

\subsection{Implementation Details}
We initialized the embedding model using the Conan-embedding-v1\footnote{\url{https://huggingface.co/TencentBAC/Conan-embedding-v1}} model \citep{li2024conanembeddinggeneraltextembedding}. We trained the model in two stages. In the first stage, we freeze the parameters of embedding model and only train the ListTransformer for 4 epochs using the training sets of cMedQA1.0, cMedQA2.0, MMarcoReranking, T2Reranking, 
huatuo\citep{li2023huatuo26m}, MARC\citep{keung-etal-2020-multilingual}, XL-sum\citep{hasan-etal-2021-xl}, and CSL\citep{li-etal-2022-csl} 
with a batch size of 1024. In the second stage, we do not freeze any parameter and use the training sets of cMedQA1.0, cMedQA2.0, and T2Reranking to train for another 2 epochs with a batch size of 256. Besides, we respectively set the hyperparameters $l$, $\gamma$, $\alpha$, and $\beta$ to 2, 10, 20, 0.2. In the first stage of training, we set $m$ to -0.2. In the second stage of training, we set $m$ to 0.1. The experiments are run on 8 NVIDIA A800 40GB GPUs.

\begin{table*}[t]
\centering
\resizebox{2 \columnwidth}{!}{
\begin{tabular}{lccccc}
\toprule
Model                                   & cMedQA1.0 & cMedQA2.0 & MMarcoReranking & T2Reranking & Avg.   \\ \hline
ListAttention                           & 90.55     & 89.38     & \textbf{43.88}           & 69.17       & \textbf{73.25} \\
Bidirectional Self-Attention                     & \textbf{90.79}    & \textbf{89.59}    & 42.12          & \textbf{69.20}      & 72.93 \\
PassageAttention & 89.73    & 89.46    & 41.08          & 69.04      & 72.33 \\
\bottomrule
\end{tabular}
}
\caption{The results of ablation study using different attentions on the reranker section of C-MTEB. All results are based on iterative inference. The best performance is in \textbf{bold}.
}
\label{tab:attention}
\end{table*}

\begin{table*}[t]
\centering
\resizebox{2 \columnwidth}{!}{
\begin{tabular}{lccccc}
\toprule
Model                                   & cMedQA1.0 & cMedQA2.0 & MMarcoReranking & T2Reranking & Avg.   \\ \hline
Fusing Original and Listwise Features          & \textbf{90.55}     & \textbf{89.38}     & \textbf{43.88}           & \textbf{69.17}       & \textbf{73.25} \\
Only Listwise Features         & 90.36    & \textbf{89.38}    & 41.48           & 68.91      & 72.53 \\
Only Original Features & 89.41    & 89.13    & 38.66          & 68.72      & 71.48 \\
Embedding Model Continue Training & 89.02    & 88.54    & 38.12          & 69.11      & 71.20 \\
\bottomrule
\end{tabular}
}
\caption{The results of ablation study using different features on the reranker section of C-MTEB. The results of Fusing Original and Listwise Features and Only Listwise Features are based on iterative inference. Iterative inference will not change the results of Only Original Feature and Embedding Continue Training, as the features of each passage are not influenced by other passages. Thus, there is no need to use iterative inference on them. The best performance is in \textbf{bold}.
}
\label{tab:features}
\end{table*}

\begin{table*}[t]
\centering
\resizebox{1.7 \columnwidth}{!}{
\begin{tabular}{lccccc}
\toprule
Model                                   & cMedQA1.0 & cMedQA2.0 & MMarcoReranking & T2Reranking & Avg.   \\ \hline
Circle Loss                           & \textbf{90.55}     & 89.38     & \textbf{43.88}           & \textbf{69.17}       & \textbf{73.25} \\
CoSENT Loss                     & 90.48    & \textbf{89.79}    & 42.64          & 68.99      & 72.98 \\
Triplet Loss & 90.13    & 88.71    & 42.04          & 68.85      & 72.43 \\ 
Cross-Entropy Loss & 89.05    & 88.95    & 35.39          & 66.65      & 70.01 \\ 
\bottomrule
\end{tabular}
}
\caption{The results of ablation study using different loss functions on the reranker section of C-MTEB. All results are based on iterative inference. The best performance is in \textbf{bold}.
}
\label{tab:loss}
\end{table*}

\subsection{Overall Results}
Table~\ref{tab:overall results} shows the results of cMedQA1.0, cMedQA2.0, MMarcoReranking, and T2Rerank-ing. Compared to other models, ListConRanker demonstrated a significant advantage, particularly on the MMarcoReranking and T2Reranking datasets. Although some previous methods are based on LLMs with larger parameter sizes (i.e., 360Zhinao-1.8B-Reranking and LdIR-Qwen2-reranker-1.5B), they use pointwise encoding, which leads to suboptimal performance. It is worth noting that we do not use the MMarcoReranking dataset in the second stage of training. The significant improvement of ListConRanker on the MMarcoReranking can be attributed to the listwise encoding introduced by the ListTransformer. Additionally, the MMarcoReranking is unique compared to the other three datasets. Each query in the MMarcoReranking corresponds to 1,000 passages. This large number of passages allows for extensive global contrastive information and interaction within the ListTransformer, resulting in a substantial performance boost.

\subsection{Ablation Study}

\subsubsection{The Effect of ListAttention}
To verify the effectiveness of ListAttention, we conducted experiments with different attention masks. The results, as shown in Table~\ref{tab:attention}, indicate that bidirectional self-attention performs worse than ListAttention. The only difference between them lies in whether the query needs to compute attention for the passages and whether the query features are influenced by the passages. This decline in performance highlights the importance of maintaining the semantic features of the query.

Additionally, we explored a PassageAttention approach. Specifically, in PassageAttention, the query has bidirectional self-attention with all passages, but each passage only computes attention between itself and the query. In this setup, to learn global comparative information, passages rely on the query as an anchor. As a result, query features are influenced by all passages, regardless of their similarity to the query. This causes the query to lose a substantial amount of its own semantic meaning. At the same time, it also reduces the efficiency of passages learning global contrastive information. Notably, PassageAttention resulted in the lowest performance, further highlighting the importance of maintaining the semantic features of query.

\subsubsection{The Effect of Fusing Original and Listwise Features}
To verify the effect of fusing listwise features with original features, we conducted experiments where we used only listwise features and only original features separately. The results, shown in Table~\ref{tab:features}, demonstrate that using only listwise features results in only a slight performance drop compared to ListConRanker. However, when the model uses only original features, there is a significant drop in performance, especially on the cMedQA1.0 and MMarcoReranking datasets. This is because the original features lack global comparative information, as they are pointwise features. This underscores that listwise features play a dominant role in ListConRanker, as well as the importance of global comparative information.

Additionally, to rule out the impact of data and training strategies, we further train the embedding model using the second-stage training strategy. We do not use the first-stage strategy, because the embedding model is frozen in the first stage. Since similarity calculation uses inner products between vectors, there is no feature interaction between the query and passage, meaning it is neither a pointwise nor a listwise model. As observed, its performance is the lowest, further emphasizing the importance of query-passage interaction in the reranker model.

\subsubsection{The Effect of Circle Loss}
To verify the effectiveness of Circle Loss, we explored the experimental results using different loss functions, as shown in Table~\ref{tab:loss}. Firstly, when using the cross-entropy loss, the model achieved the worst performance due to inefficient data sampling. Secondly, when the loss function was switched to Triplet Loss \citep{balntas2016learning} and CoSENT Loss, which are designed specifically for ranking tasks, the models are able to sample all positive and negative samples of query in each training step. This not only improved the data sampling efficiency but also enabled the ListTransformer to bring positive samples closer together, which obtains more comprehensive global contrastive information. Finally, when the model used Circle Loss, compared to Triplet Loss and CoSENT Loss, the model has smoother gradient changes during training. Specifically, Circle Loss increases the gradient when the model is far from the optimal space and reduces the gradient as it approaches the optimal space. This allows the model to find the optimal space more easily, leading to the best performance.

\subsubsection{The Effect of Iterative Inference}
To explore the effect of iterative inference, we compared the results of ListConRanker using iterative inference with those using non-iterative inference. The results are shown in Table~\ref{tab:overall results}. In the non-iterative inference approach, we input all passages into the ListConRanker at once. And we directly sort the passages based on their output scores to obtain all the results. It can be observed that iterative inference improves performance on both the cMedQA1.0 and MMarcoReranking datasets, especially on MMarcoReranking. This is due to the fact that each query in MMarcoReranking corresponds to a large number of candidate passages, as mentioned above. In contrast, iterative inference has a slight negative effect on cMedQA2.0. This might be because all the negative passages in the cMedQA2.0 are very similar to the query, causing the positive passages to be ranked lower in the early iterations. Further, this prevents them from being input into the final iteration.

\subsubsection{The Effect of Scaling Up ListTransformer}
\begin{figure}[t!]
    \centering
    \includegraphics[width=\linewidth]{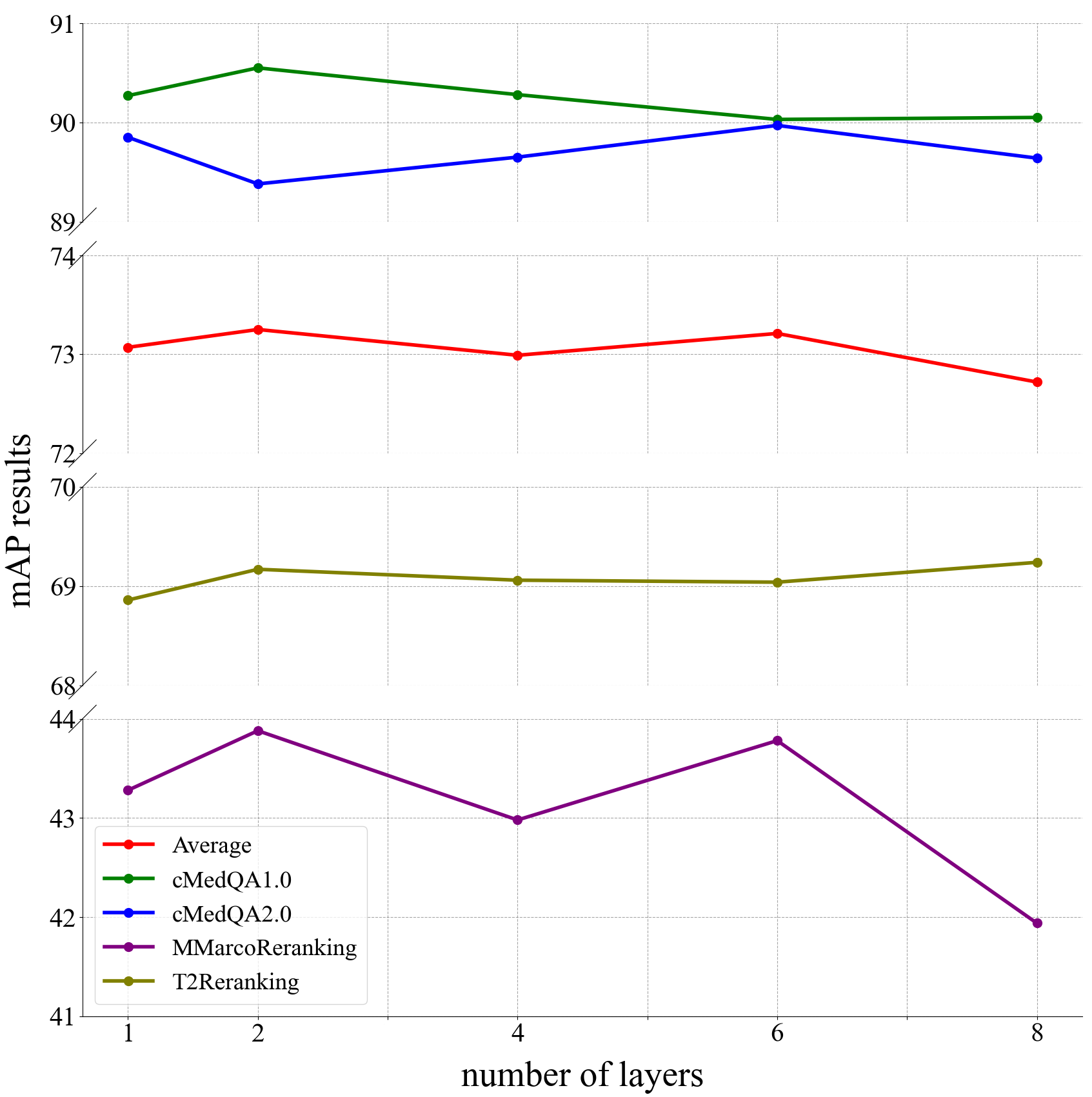}
    \caption{
    The influence of different layers of ListTransformer.
    }
    \label{fig:layer}
\end{figure}

To explore whether ListConRanker has a scaling-up ability similar to LLMs \citep{kaplan2020scaling}, we conducted experiments with different numbers of ListTransformer layers. And the results are shown in Figure~\ref{fig:layer}. However, as the number of layers increased, we do not observe a significant improvement in the performance of ListConRanker. In most cases, the number of passages corresponding to a query is small. Therefore, a smaller number of ListTransformer layers is already sufficient for effective comparison among passages and learning global comparative information. Moreover, this also demonstrates the robustness of ListConRanker to the layer of ListTransformer. ListConRanker outperformed the state-of-the-art model across all layer settings except for the 8-layer configuration, which might be due to the lack of an increase in the training data size alongside the training parameters. This caused the model to be undertrained.

\section{Conclusion}
In this paper, we propose a novel reranker named ListConRanker by introducing listwise encoding through the use of the ListTransformer. The listwise encoding can help learn the global contrastive information between passages. In addition, we use ListAttention to help maintain the features of the query, which assist similarity calculations with passages. Finally, we propose to use the circle loss to replace the cross-entropy loss and solve the problem of data efficiency. Besides, circle loss can smooth the gradient and help the model to find the optimal space. The experimental results on the reranking benchmark of C-MTEB demonstrate the effectiveness of ListConRanker.

\section{Limitations}
We propose using iterative inference to address the issue of disperse attention when inputting a large number of passages, especially in the MMarcoReranking dataset. This problem arises due to the lack of training samples with large numbers of passages. However, iterative inference leads to the need for multiple inferences on some passages, which increases time complexity. For datasets where each query corresponds to only a small number of passages, the impact of the increased time complexity is subtle. We have other methods to address the issue of large numbers of passages while reducing time complexity. For example, we can sample a small number of passages from the passage set $P$. We remove these samples from $P$ after obtaining the similarity scores. And we repeat the process multiple times to obtain all similarity scores and get the final order. However, the randomness of sampling can cause instability and uncertainty in the results. Therefore, we need to explore a low-time-complexity inference method for simultaneously inputting a large number of passages in the future.

% \section*{Acknowledgments}

\bibliography{custom, anthology}

% \appendix

% \section{Example Appendix}
% \label{sec:appendix}

% This is an appendix.

\end{document}